# A Survey of TinyML Applications in Beekeeping for Hive Monitoring and Management

WILLY SUCIPTO, University of Technology Sydney, Australia

JIANLONG ZHOU, University of Technology Sydney, Australia

RAY SEUNG MIN KWON, BeeSTAR (powered by LB Agtech Holdings Pty Ltd), Australia

FANG CHEN, University of Technology Sydney, Australia

Honey bee colonies are essential for global food security and ecosystem stability, yet they face escalating threats from pests, diseases, and environmental stressors. Traditional hive inspections are labor-intensive and disruptive, while cloud-based monitoring solutions remain impractical for remote or resource-limited apiaries. Recent advances in Internet of Things (IoT) and Tiny Machine Learning (TinyML) enable low-power, real-time monitoring directly on edge devices, offering scalable and non-invasive alternatives. This survey synthesizes current innovations at the intersection of TinyML and apiculture, organized around four key functional areas: monitoring hive conditions, recognizing bee behaviors, detecting pests and diseases, and forecasting swarming events. We further examine supporting resources, including publicly available datasets, lightweight model architectures optimized for embedded deployment, and benchmarking strategies tailored to field constraints. Critical limitations such as data scarcity, generalization challenges, and deployment barriers in off-grid environments are highlighted, alongside emerging opportunities in ultra-efficient inference pipelines, adaptive edge learning, and dataset standardization. By consolidating research and engineering practices, this work provides a foundation for scalable, AI-driven, and ecologically informed monitoring systems to support sustainable pollinator management.

## 1 Introduction

The honey bee (*Apis mellifera*) plays a crucial role in human civilization, contributing not only to biodiversity but also to various industries [51]. These bees produce a wide range of valuable products, including honey, bee pollen, candles, cosmetics, propolis, and mead. More importantly, nearly one-third of global food production depends on pollination by honey bees [127]. Despite

Authors' Contact Information: Willy Sucipto, willy.sucipto@student.uts.edu.au, University of Technology Sydney, Sydney, Australia; Jianlong Zhou, University of Technology Sydney, Sydney, Australia, Jianlong.Zhou@uts.edu.au; Ray Seung Min Kwon, BeeSTAR (powered by LB Agtech Holdings Pty Ltd), Sydney, Australia, ray@beestar.com.au; Fang Chen, University of Technology Sydney, Sydney, Australia, Fang.Chen@uts.edu.au.

their ecological significance in maintaining environmental balance [35], bee colonies worldwide are increasingly threatened by various threats [131]. These challenges include parasites, colony collapse disorder (CCD), environmental changes, and swarming, all of which negatively impact health and productivity [15, 76, 87].

Given these threats, it is of the utmost importance for beekeepers to monitor and manage beehive conditions continuously [139]. However, many beekeepers still rely on manual observation techniques to assess colony conditions and apply treatments [4]. This conventional approach has several drawbacks: (1) direct hive inspections can disturb the bees and induce stress [7], (2) handling the hive frames may accidentally harm or kill bees [110], and (3) manual monitoring is time-consuming and inefficient. These limitations are particularly evident during the winter season, when beekeepers are restricted to observing hives externally without directly assessing colony health [32]. Given the importance of timely intervention in beekeeping, there is a pressing need for continuous, minimally invasive, and automated hive monitoring techniques that reduce human intervention while supporting colony well-being [128].

Recent technological advancements offer new opportunities for modern beekeeping by enabling precise monitoring and management of bee health with minimal human interference [34]. These innovations leverage state-of-the-art technologies such as the Internet of Things (IoT) and Machine Learning (ML). The integration of IoT and ML into small, low-power computing devices is referred to as Tiny Machine Learning (TinyML). The term "Tiny" signifies its compact size and ultra-low power consumption [32], yet it could perform essential ML tasks. A key advantage of TinyML is its suitability for deployment in remote and rural areas, where access to stable internet connectivity and power infrastructure is limited. By processing data locally on edge devices, TinyML eliminates the need for constant cloud communication, enabling real-time monitoring and decision-making even in isolated beekeeping locations. This makes it an ideal solution for rural apiculture, supporting sustainable practices and empowering beekeepers with actionable insights in areas where traditional digital technologies are impractical.

In the context of beekeeping, TinyML can seamlessly integrate into hive monitoring systems, transforming conventional methods into real-time, data-driven decision-making processes [45]. The IoT component consists of integrated sensors, such as temperature, humidity, weight, acoustic, and camera sensors, which continuously monitor hive conditions [85, 102]. The collected data is then processed on microcontrollers or edge computing devices, enabling real-time insights without reliance on cloud computing [7]. This localized processing is especially valuable in rural and remote beekeeping settings, where internet access is limited or unreliable. By functioning independently of cloud infrastructure, TinyML ensures continuous monitoring and timely alerts, even in isolated environments.

On the ML side, TinyML utilizes specialized ML models to analyze hive data. For example, Convolutional Neural Networks (CNNs) can detect the presence of bees in images and classify their health status [19]. Similarly, audio-based ML models can predict the presence of a queen bee by analyzing hive sound patterns [129]. These advancements enable beekeepers to make informed decisions, enhancing colony management while minimizing disturbances.

The integration of technological advancements in beekeeping introduces complex interactions between modern monitoring tools and the natural behaviors of bee colonies. While these innovations offer new opportunities, they also raise concerns about their long-term impact on bee populations and the surrounding environment [126]. Understanding these interactions requires a comprehensive review that integrates both ecological perspectives and technological innovations, ensuring that advancements in beekeeping support sustainability and colony well-being [142].

Current research often focuses on isolated perspectives, emphasizing either biological aspects of bee health or the technological capabilities of monitoring systems, while rarely examining their interplay [140]. A holistic approach is essential to bridge this gap, particularly by integrating TinyML-driven solutions that maintain ecological balance [98]. This review aims to provide such a perspective, offering a structured review of TinyML applications in beekeeping and their role in establishing sustainable and productive hive management practices.

The purpose of this review is to examine the recent research, covering a vast area of findings that integrate ecological insight into beehive life and the technologies applied in beekeeping. There are three primary objectives of this study: (1) to offer a comprehensive perspective that bridges ecological and technological domains; (2) to connect the critical gaps in the existing research; and (3) to explore promising areas of development in the future. For example, while conventional mite detection methods involve destructive techniques such as alcohol washes, recent innovations leverage computer vision techniques for non-invasive monitoring within the hive. Bridging such gaps can help stakeholders with the utility or knowledge to keep ecological balance by developing potential tools and contributing to decision-making in this field or related areas. The main contributions of this research are summarized as follows:

- A structured synthesis of IoT and TinyML-based technologies for addressing apicultural challenges, with emphasis on lightweight and edge-deployable methods.
- A comprehensive review of recent ML and TinyML applications for non-invasive hive monitoring and management, including hive activity recognition, disease and pest detection, and swarm prediction.
- The identification of current dataset and benchmarking limitations, alongside research gaps and future directions for deploying IoT and TinyML in scalable, ecologically responsible ways.

The remainder of this work is organized as follows: Section 2 provides background on TinyML and IoT in beekeeping, including ecological challenges and the role of technology in addressing them. Section 3 focuses on hive monitoring as the core TinyML application in beekeeping and details its key components, including bee activity recognition, pest and disease detection, and swarming and anomaly prediction. Section 4 reviews existing datasets and benchmarking practices relevant to TinyML in this domain. Section 5 discusses current challenges and future directions, including hardware constraints, model optimization trade-offs, data issues, and deployment considerations. Section 6 presents a broader discussion on the ecological impact of these technologies, technological benefits, stakeholder implications, and ethical considerations. Finally, Section 7 concludes the study and outlines recommendations for future research.

## 2 Background on TinyML and IoT in Beekeeping

### 2.1 Ecological Challenges in Beekeeping

As outlined in Section 1, honey bees (Apis mellifera) are vital to global biodiversity, food security, and agricultural productivity [127]. Pollination by honey bees is essential for many crops, contributing substantially to agricultural productivity and ecological balance [35]. In addition to their vital ecological role, honey bees also produce commercially valuable hive products such as honey, beeswax, and other apicultural products [70, 91]. Despite their importance, bee colonies worldwide are continually affected by an increasing number of threats [76, 131, 139]. These challenges include environmental and anthropogenic stressors such as habitat degradation [55, 65, 109], climate change [90, 132], pesticide exposure, and pathogen proliferation, all of which diminish bee health and productivity.

These environmental and anthropogenic stressors collectively weaken bee colonies, increasing their susceptibility to colony losses [123]. In the most severe scenarios, they contribute to phenomena such as CCD, where worker bees suddenly vanish from a colony [38, 39, 130], or excessive swarming, which can reduce colony stability and workforce availability [133]. If CCD and excessive swarming are not addressed promptly, they could have far-reaching consequences, impacting the environment, food industry, and global economy [36].

The interaction among multiple stressors often exacerbates colony instability, as these factors rarely occur in isolation [54]. For instance, pesticide exposure may compromise bees' immune systems, increasing their vulnerability to parasites like *Varroa destructor* [41, 100], while habitat loss driven by climate change further limits foraging resources [139]. Given the complexity of these challenges, traditional hive inspection methods are increasingly insufficient, especially when it comes to detecting subtle yet critical changes in environmental factors such as temperature and humidity, which are crucial to brood development, foraging behavior, and overall colony survival [6]. To address these limitations, advanced technological solutions, such as IoT-based monitoring and TinyML-driven predictive models, offer a promising approach to early anomaly detection, risk assessment, and proactive intervention, potentially improving colony resilience and sustainability [4, 9].

### 2.2 Role of IoT and ML in Addressing Beekeeping Challenges

Given the complexity of these interrelated stressors, it is critical to develop improved monitoring methods that are both scalable and minimally invasive [86]. As shown in Figure 1, traditional hive inspections remain an essential practice for assessing colony health, but they are labor-intensive, infrequent, and often inadequate for detecting early signs of stress or disease. To overcome these limitations, researchers have proposed integrating IoT-based systems and data-driven platforms to enable real-time, non-intrusive hive monitoring and improve the timeliness and accuracy of colony assessments [45]. Recent advancements in sensor technologies and Artificial Intelligence (AI) now support predictive analytics, proactive risk management, and informed decision-making with minimal human input [126]. These innovations significantly enhance beekeeping efficiency, enable precise hive management, and improve responses to complex ecological and anthropogenic challenges [79].

Recent developments in IoT technologies have enabled the deployment of low-power sensors to continuously monitor hive and environmental conditions with minimal human intervention [71]. Commonly deployed sensors include measuring internal hive temperature and humidity, weight, sound frequencies, $CO_2$ levels, and visual data through cameras [93]. In addition to in-hive sensing, miniature bee-mounted sensors have also been utilized to infer colony-level pollination activity, offering a complementary perspective by capturing direct behavioral patterns in the field [3]. These platforms typically consist of microcontrollers, inertial measurement units (IMUs), and batteries, and are carefully designed to minimize weight and avoid impairing flight. While only a limited number of bees (typically 10 to 15 per trial) are equipped in each experiment, the data collected provides valuable insights into colony-wide behavior and landscape-level pollination dynamics. Together, these energy-efficient IoT systems support continuous data acquisition, enabling early anomaly detection and responsive hive management [49, 85].

To transform raw sensor data into meaningful insights, ML algorithms are increasingly integrated into IoT-based hive monitoring systems. These models detect early behavioral and physiological signs that are difficult to uncover through manual inspection. Supervised models such as support vector machines (SVM), random forests, and CNN have been used to classify hive conditions and detect behavioral anomalies [10, 89, 134]. They typically rely on data from cameras, audio, temperature, and other environment sensors to predict colony-level events such as swarming, brood disruption, or queen loss [4, 9]. For instance, acoustic models have been applied to detect stress-related buzzing, swarming indicators, or the absence of a queen [52, 59], while vision-based models have identified brood stages, comb health, or pest intrusion [34, 97]. These ML systems can operate either in the cloud for complex processing or directly on-site, which is particularly useful for deployment in remote or rural areas.

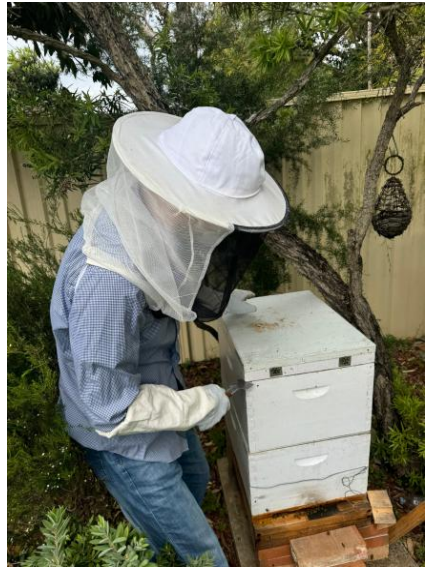

Fig. 1. Manual hive inspections are essential but limited in detecting early signs of stress, underscoring the need for TinyML-based monitoring solutions

While conventional ML models offer valuable predictive capabilities, they often depend on high-performance hardware or cloud connectivity, which can be impractical in field conditions. TinyML addresses this challenge by enabling lightweight inference directly on low-power, resource-constrained devices [94, 124]. Techniques such as model quantization, pruning, and knowledge distillation are often applied to reduce model size and memory usage (flash and RAM) while preserving sufficient accuracy for real-time inference [13, 136]. In beekeeping, TinyML enables continuous, on-device analysis of environmental and behavioral data without the need for persistent internet connectivity or cloud processing [62]. This is especially beneficial in remote or rural apiaries, where connectivity and power supply are limited. As illustrated in Figure 2, the system integrates in-hive sensor nodes with embedded ML to perform real-time inference locally. When necessary, selected outputs can be transmitted via low-power networks for external storage or beekeeper insight, supporting the development of autonomous, scalable, and minimally invasive monitoring systems [4, 45].

## 2.3 Nano-enabled IoT Sensors for High-Resolution Hive Monitoring

Recent advances in nano-enabled sensor technology offer promising enhancements to traditional IoT-based beekeeping systems. Conventional sensors, such as those measuring temperature, humidity, or hive weight, provide valuable metrics but often fall short in detecting early physiological and environmental stress indicators within the hive. To address this, nanomaterials such as metal-oxide semiconductors (MOS), carbon nanotubes (CNT), and functionalized nanocomposites have been integrated into gas-sensing platforms due to their high surface area, chemical specificity, and sensitivity at parts-per-billion (ppb) concentrations [25, 95].

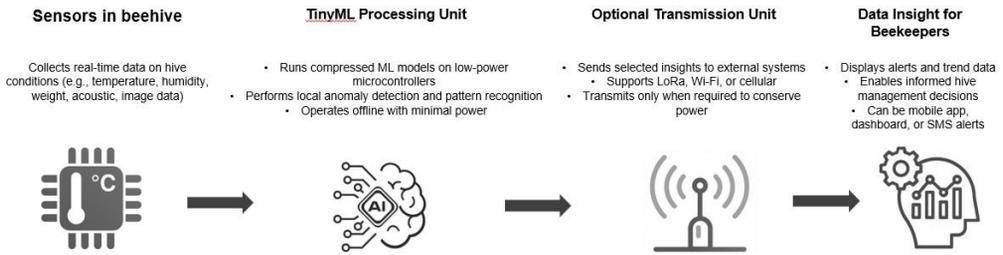

Fig. 2. System overview of TinyML-enabled hive monitoring, integrating local inference from in-hive sensors with optional low-power data transmission to support real-time beekeeper insights.

These nano-sensors are capable of detecting volatile organic compounds (VOCs) that signal subtle hive disturbances, including queen pheromone imbalances, pathogen activity, or poor ventilation. Recent studies have highlighted the diagnostic potential of VOC ratios, particularly the ethyl palmitate to ethyl oleate ratio, in detecting early-stage Varroa mite infestations [143]. Monitoring the relative concentrations of key volatile organic compounds (VOCs) offers a promising method for early detection of stressors such as Varroa mite infestation or declining queen vitality [17]. Leveraging these chemical indicators in nano-sensor platforms enables non-invasive, high-resolution monitoring of colony health in real time.

Building on these advances, nano-gas sensor systems have been increasingly integrated with edge AI platforms to enable localized, real-time analysis of hive air. These embedded systems are designed for low-power operation and can be deployed in remote or off-grid apiaries without requiring constant internet connectivity. By conducting on-device inference, they support autonomous, non-intrusive monitoring that scales well across multiple sites [5]. Beyond immediate detection, the high-resolution phenotypic data generated from these platforms can inform selective breeding strategies and strengthen biosecurity efforts by identifying colony stress trends across diverse environments [51, 57].

## 3 Hive Monitoring: TinyML Applications in Beekeeping

Hive monitoring refers to systematic observation and measurement of internal and external hive parameters to assess colony health, productivity, and early signs of stress or disease. With the advancement of embedded intelligence, TinyML enables low-power devices to continuously track and analyze diverse data streams such as temperature, humidity, hive weight, bee activity, acoustic signals, and visual cues directly within the hive. Table 1 summarizes representative TinyML-based hive monitoring systems, highlighting key sensors, deployed models, and intended monitoring objectives [66, 67].

To overcome the limitations of traditional manual monitoring, TinyML-based solutions enable embedded microcontrollers to process sensory data locally, identifying early indicators of stress or imbalance without relying on cloud resources [2]. These systems autonomously detect deviations in hive metrics such as temperature, humidity, $CO_2$, and hive weight that may signal disruption, brood neglect, or environmental interference [85, 140].

The primary objective of hive condition monitoring is to ensure environmental stability within the colony by tracking physiological or behavioral deviations [103, 104]. TinyML systems offer reliable, autonomous, and energy-efficient monitoring suitable for long-term deployment in remote or off-grid locations. Studies by Abdollahi et al. [4] and Lin et al. [62] have demonstrated the feasibility of on-device anomaly detection and time-series classification in real-world hive environments. This section reviews three critical application areas of TinyML in beekeeping [125]:

- Bee Behavior and Activity Recognition [46, 71]
- Pest and Disease Detection [74]
- Swarming and Anomaly Prediction [64]

Table 1. TinyML-based hive condition monitoring systems

| Sensors | Example Models | Purpose | References |
|---|---|---|---|
| Temperature | DHTXX, SHT3X | Monitors internal hive temperature to ensure optimal brood development and overall hive health. | [23, 53, 77] |
| Humidity | SHT3X, AM2320 | Tracks humidity levels to maintain proper environmental conditions and avoid stress on bees. | [53, 77] |
| Weight | HX711 Load Cell | Measures hive weight to monitor honey production, food stores, and overall colony size. | [73, 138] |
| Acoustic | MEMS Mic, Piezo-electric | Detects bee sounds and vibrations to analyze activity levels, stress indicators, and swarming behavior. | [23, 84] |
| Gas | MQX, NDIR | Monitors $CO_2$ levels for hive ventilation and air quality, indicating colony metabolism and health. | [23, 120] |
| Optical | Infrared | Tracks bee entry and exit patterns, which help gauge foraging activity and colony strength. | [21, 27] |
| Camera/Visual | Thermal Cam, Standard Cam | Captures images or videos to visually monitor hive activity, brood health, and pests like Varroa mites. | [33, 80] |
| Vibration | Accelerometer, Piezoelectric | Detects hive vibrations to assess colony activity and detect potential threats or anomalies. | [129] |

## 3.1 Bee Activity Recognition

Bee activity recognition involves detecting and interpreting behavioral patterns that reflect the colony's health and functional dynamics. Understanding these behaviors allows beekeepers to assess colony strength, productivity, reproductive status, and responses to environmental stressors[46, 90]. This includes behaviors such as foraging, guarding, waggle dancing, fanning, and brood care. Recognition typically relies on three core sensing modalities: acoustic signals, visual tracking, and vibrational/motion sensing [61].

Among these, acoustic monitoring captures hive sounds associated with specific bee behaviors, such as fanning or buzzing related to stress or queen presence [4, 29]. These signals are processed using features like Mel spectrograms or MFCCs and analyzed by lightweight CNNs or RNNs [61]. This method is particularly effective for identifying behavior transitions, detecting queen presence, or predicting swarming [104]. Expanding on this, Kim et al. [52] proposed an acoustic scene classification framework using MFCCs, mel spectrograms, and constant-Q transforms, comparing machine learning models (SVM, Random Forest, XGBoost) with deep CNNs. Their VGG-13 model achieved 91.93% accuracy and, with Grad-CAM, provided interpretable insights by highlighting key frequency-time regions, underscoring the potential for CNN-based systems to detect abnormal hive conditions. Similarly, Kulyukin et al. [59] demonstrated that a shallow convolutional neural network (RawConvNet) trained directly on raw audio achieved 99.93% accuracy on BUZZ1 and 96.53% on BUZZ2 datasets, outperforming both deeper CNNs and traditional ML models. Their system validates the practicality of TinyML-based audio monitoring for real-time, in-hive analysis without requiring cloud computation. In line with this direction, Soares et al. [112] introduced a compact MFCC-based descriptor combining cepstral, time, and frequency features to detect queen presence. Using only 31 selected MFCCs and a Random Forest Classifier, their approach achieved 99% accuracy while maintaining low computational requirements, highlighting the feasibility of accurate, low-resource queen detection on embedded systems.

In parallel, vision-based monitoring employs camera systems and computer vision algorithms to observe bee behaviors at the hive entrance or within the hive. These systems can detect and classify key activities, such as foraging departures, pollen load detection, guarding, or swarming cues. For instance, Bilik et al.[18] evaluated motion-based analysis, CNN classification, and YOLOv8 detection, finding that ResNet-50 achieved the highest accuracy (98%) while smaller models like YOLOv8n showed potential for TinyML deployment. Similarly, Ratnayake et al.[82] proposed a hybrid detection and tracking system (HyDaT) combining background subtraction with YOLOv2 to monitor unmarked bees in natural floral environments, achieving 86.6% detection with reduced processing cost. Ngo et al. [77] developed a YOLOv3-tiny-based imaging system deployed on Jetson TX2 to distinguish pollen-bearing bees and assess foraging behavior, achieving an F1-score of 0.94 and operating continuously over 112 days. These approaches demonstrate that lightweight object detection and tracking models can effectively capture behavioral patterns relevant to colony health. While visual systems offer detailed insights, they face challenges with occlusion, lighting conditions, and computational cost, though recent work on edge-based deployment increasingly addresses these barriers in the context of TinyML and scalable hive monitoring.

Complementing acoustic and visual modalities, vibrational and motion-based sensing provides a valuable non-invasive method for monitoring internal hive dynamics. These systems typically utilize accelerometers, piezoelectric sensors, or laser vibrometers to capture subtle substrate-borne signals that reflect bee movement and collective activity. Uthoff et al. [129] demonstrated that vibrational sensing could effectively distinguish between queenright and queenless colonies, as well as detect pre-swarming behavior, based on frequency differences in comb vibrations. Their study emphasized that vibration signals are often less affected by environmental noise than acoustic ones, making them particularly robust for in-hive analysis. Similarly, Zgank [141] used vibration recordings from hive frames combined with machine learning models, particularly decision trees and random forests, to classify swarm-related activity. The best-performing model achieved 97.14% accuracy using entropy, zero-crossing rate, and energy-based features extracted from time-domain signals. These findings support the viability of low-power, vibration-based monitoring systems in precision apiculture, especially when paired with efficient TinyML-compatible models.

Each sensing modality, including acoustic, visual, and vibrational, offers distinct advantages and faces unique challenges when applied to bee activity recognition. Acoustic sensing provides low-power, internal monitoring and is particularly effective for detecting stress behaviors and queen presence, while visual systems excel in capturing entrance activity, pollen foraging, and external threats, albeit with higher computational demands. Vibrational and motion-based sensors, on the other hand, enable continuous, non-intrusive monitoring of internal dynamics and have shown resilience in noisy or light-variable environments. In practice, combining multiple modalities

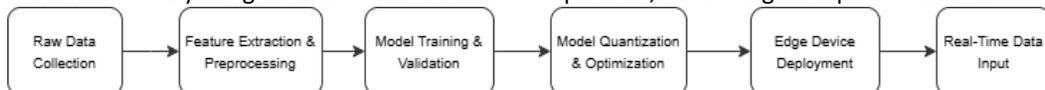

Fig. 3. TinyML workflow for hive monitoring applications, illustrating key stages from data preparation and model training to model optimization and deployment for real-time inference on embedded edge devices

may improve robustness, increase behavioral coverage, and compensate for limitations in any single system. Recent research is increasingly exploring such multimodal approaches as a path toward more reliable and scalable hive intelligence platforms.

Figure 3 illustrates the end-to-end TinyML workflow commonly applied in hive monitoring systems. The process begins with raw data collection from acoustic, visual, or vibrational sensors, followed by feature extraction and preprocessing tailored to the specific sensing modality. Machine learning models are then trained and validated using this processed data before undergoing quantization and optimization steps to reduce memory footprint and improve energy efficiency, essential for resource-constrained edge deployment [4, 64]. Once optimized, the models are deployed to embedded devices such as microcontrollers or edge accelerators within or near the hive [29]. These models then operate in real-time, continuously processing incoming sensor data to infer specific bee behaviors or colony conditions. This workflow enables low-latency, on-device inference, aligning
with the goals of scalable, non-invasive, and energy-efficient monitoring in precision apiculture.

## 3.2    Pest and Disease Detection

Early and accurate detection of pests and diseases is crucial to maintaining honey bee colony health and ensuring the productivity of apiaries. Invasive mites such as *Varroa destructor*, pathogens like *Nosema ceranae*, and predatory hornets such as *Vespa velutina* are among the primary biological threats contributing to colony collapse and population decline [56, 69]. Traditional detection methods rely on manual inspections and lab-based diagnostics, which are labor-intensive, infrequent, and often miss early-stage infections [22, 38, 122]. As an alternative, emerging TinyML-enabled systems offer automated, non-invasive detection of these threats within or around the hive [47, 72].

Recent approaches leverage acoustic, environmental, visual, and chemical signals to detect signs of infection or infestation in honey bee colonies [17, 64, 115]. Audio signals captured via embedded microphones can reveal anomalies in bee behavior linked to stress or parasitic activity [42]. These are analyzed using Mel spectrograms and lightweight models such as CNNs or LSTMs [11, 121]. Environmental sensors measuring temperature, humidity, and hive weight also provide valuable cues; for example, shifts in thermal patterns or foraging mass may signal brood issues or pathogen presence [140]. Complementing these approaches, chemical detection methods utilizing gas sensors have demonstrated the ability to identify volatile organic compounds (VOCs) that serve as earlystage biomarkers for diseases such as American foulbrood or Varroa infestations [115, 137]. Together, these multi-modal sensing strategies enable a more comprehensive understanding of hive health while supporting the design of real-time, low-power diagnostic systems.

Visual sensing, in particular, has shown strong potential for both internal hive monitoring and external threat detection. Internally, lightweight CNN-based classification models are being developed to distinguish between healthy bees, pollen carriers, and individuals infected with Varroa destructor using image data captured at the hive entrance [18, 37]. Torky et al.[121] implemented a MobileNet-based deep learning pipeline that achieved 95% classification accuracy across six bee health conditions, including Varroa infection, queen absence, and ant infestations, by optimizing CNN layers for embedded deployment. As illustrated in Fig.4, such models enable automated, on-device identification of individual bee health status using image-based object detection and classification. In addition to internal health assessment, vision-based techniques are also being used to detect external biological threats. Nasir et al. [75] proposed a multi-modal recognition framework combining low-resolution infrared imagery with 3D flight trajectories to identify *Vespa velutina* and *Vespa orientalis* hornets in unconstrained outdoor conditions. By integrating trajectory-derived features with convolutional models such as GoogLeNet and Xception, their system achieved 97.1% accuracy under real-world conditions. These studies collectively demonstrate the viability of visual pipelines in TinyML applications, supporting early warning systems for both in-hive abnormalities and predatory incursions.

Environmental and chemical sensing have gained traction as non-invasive alternatives for detecting microbial infections and parasite infestations within hives. Volatile organic compounds (VOCs), emitted as metabolic byproducts by bees or pathogens, serve as promising biomarkers for early disease diagnosis. Bikaun et al.[17] identified a distinct chemical profile in hives infected with Paenibacillus larvae, the agent behind American foulbrood (AFB), with elevated levels of compounds such as phenol and 2-nonanone. This study demonstrated that VOC analysis could detect AFB prior to the onset of visible symptoms, enabling proactive colony management. Similarly, Szczurek et al.[115] validated the use of portable gas sensors to identify Varroa destructor infestations, revealing statistically significant differences in sensor responses corresponding to mite load. Wilson [137] further contextualizes these findings by reviewing recent applications of electronic nose (e-nose) technology across human and veterinary domains, highlighting its potential for real-time, low-power, disease-specific detection in embedded systems. These works establish VOC sensing as a viable pathway for TinyML-based diagnostics through low-power signal acquisition and pattern recognition.

Beyond single-modality sensing, recent frameworks have demonstrated the advantages of integrating heterogeneous data streams for robust anomaly detection in hives. Zaman and Dorin [140] proposed the Operational–Investigative–Predictive (OIP) framework to classify hive monitoring systems and emphasize the need for predictive models grounded in multi-sensor integration. Machhamer et al.[64] advanced this by implementing a modular IoT system with visual programming to detect hive anomalies using SVM-based classifiers across multiple sensor streams. Expanding further, Rathore and Agrawal[96] demonstrated an automated deep learning pipeline for brood stage classification using enhanced Inception-v3 models. Their method achieved over 97% detection accuracy and 94.93% classification accuracy for eggs, larvae, and pupae using image segmentation, circular Hough transform, and CNN-based classification, showing strong potential for real-time hive diagnostics with embedded deployment options. Similarly, Anwar et al. [11] introduced Apis-Prime, a compact CNN-LSTM model designed to predict daily hive weight using historical sensor data. With fewer than 0.5 million parameters and sub-10 ms inference latency, the model exemplifies how lightweight temporal architectures can support predictive hive analytics in energy-constrained environments.

Despite promising advances, several challenges persist in the practical deployment of TinyMLbased pest and disease detection systems. Sensor drift, environmental noise, and variability across hive configurations can affect model generalization and detection accuracy. In particular, the lack of standardized datasets for rare conditions like early-stage infestations or colony infections hinders the performance of supervised learning methods. Research is increasingly exploring few-shot learning, continual learning, and hybrid edge–cloud strategies to improve model adaptability in dynamic apiary environments [62, 80]. As sensing modalities and ML architectures continue to evolve, future efforts should emphasize energy-efficient design, context-aware feature selection, and low-latency inference pipelines. These advancements will not only enhance health diagnostics but also enable the prediction of broader behavioral phenomena such as colony swarming, which is closely tied to stress indicators and internal hive dynamics, an area explored in the following subsection.

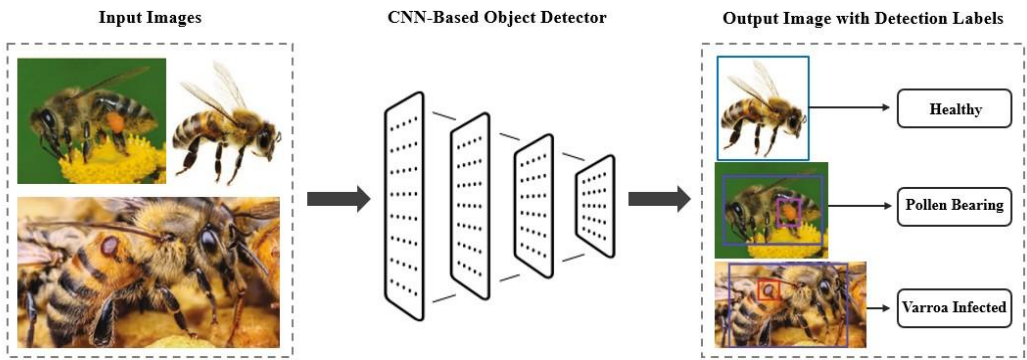

Fig. 4. A CNN-based image classification pipeline used for identifying bee health status. The model takes bee images as input and classifies them into categories such as pollen-bearing, Varroa-infected, or healthy, supporting automated monitoring and early detection of threats in the hive

### 3.3 Swarm Prediction

Swarming is a natural reproductive process in honey bee colonies, wherein the old queen departs with a substantial portion of the worker population to establish a new colony. While this phenomenon is ecologically significant, it poses major challenges for beekeepers due to the associated decline in hive productivity and population [36]. Accurate and timely swarm prediction remains difficult, as conventional methods rely on manual detection of queen cells or behavioral cues, which are often labor-intensive and unreliable. Recent studies have proposed sensor-based approaches to automate this process. Bencsik et al.[16] demonstrated that hive vibrations, monitored through accelerometers and analyzed using principal component analysis (PCA), can reveal distinct precursors to swarming several days in advance. Their work highlights the potential of vibrational signatures as non-invasive indicators of colony-level transitions. Complementing this, Ngo et al.[77] employed a lightweight deep learning-based imaging system on a Jetson TX2 to track foraging dynamics, showing that fluctuations in pollen foraging activity are tightly linked to environmental and colony-level factors. Although not explicitly designed for swarming, such behavioral indices could contribute to multi-modal swarm prediction frameworks. Furthermore, Robustillo et al. [105] emphasized the role of highly correlated time-series sensor data, such as internal temperature, humidity, and weight, and proposed multivariate forecasting models that could help detect anomalies consistent with swarm preparation. Together, these approaches lay the foundation for predictive, embedded swarm monitoring using TinyML-compatible architectures.

Recent innovations have explored diverse sensing modalities to detect swarming activity through visual, acoustic, and vibrational signals. Voudiotis et al.[133] developed a smart in-hive monitoring system using embedded imaging and lightweight CNNs to classify bee clustering patterns into five distinct swarming stages, allowing automated alerts via a mobile interface. Similarly, Kim et al.[52] applied convolutional neural networks to acoustic features such as MFCCs and mel spectrograms, achieving over 90% classification accuracy and leveraging Grad-CAM for explainable insights into sound-based hive conditions. Complementing these approaches, Bencsik et al. [16] demonstrated that hive vibration signals, collected via wall-mounted accelerometers and analyzed with principal component analysis (PCA), can reveal a time-evolving vibrational signature that anticipates swarming several days in advance. Their method captures statistically independent spectral components that exhibit characteristic divergence and polarity reversal patterns prior to a swarm event, offering a robust, non-invasive prediction framework well suited for embedded deployment. Together, these studies demonstrate the viability of multimodal swarm detection systems deployable on edge devices using TinyML. Expanding on this, Krzywoszyja et al. [58] proposed a statistical distribution-based approach for swarm detection, where sound and vibration samples were analysed using kernel density estimation (KDE) and compared with known swarm patterns through the Kolmogorov–Smirnov test. Their method successfully identified swarm events in real recordings and was later implemented on a prototype embedded system, reducing computational complexity by 86% while maintaining detection reliability. The system achieved early detection of swarm activity up to 33 minutes before the culmination of the event, demonstrating both predictive capability and feasibility for low-power deployment.

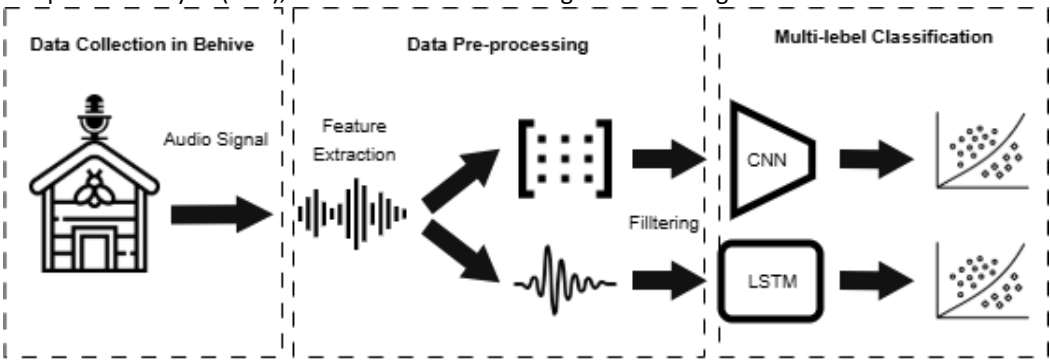

Fig. 5. Pipeline for swarm behavior detection using acoustic data: audio signals are collected via microphones, processed through feature extraction and filtering, and analyzed by CNN or LSTM-based classifiers for multi-label classification of hive activity

A representative architecture for acoustic-based swarm prediction is shown in Fig.5. In a typical workflow, hive sounds are transformed into time–frequency representations such as Mel spectrograms or MFCCs, preprocessed to reduce background noise, and then classified using lightweight CNNs or LSTMs. Recent studies demonstrate the potential of this approach: Iqbal et al.[43] evaluated multiple machine learning (Naïve Bayes, KNN, SVM) and deep learning (CNN, LSTM, Transformer) models on the NU-hive dataset, which contains over 47 GB of hive audio recordings, and reported that SVM with Mel spectrograms achieved 97% accuracy while CNN with MFCCs reached 99%. Building on this structure, Várkonyi et al. [116] introduced a dynamic noise filtering method that significantly improved robustness under real-world conditions such as wind and rain, enabling more reliable classification of hive states, including foraging, fanning, silence, and pre-swarming activity. Together, these advances underscore the strong discriminative power of acoustic features for swarm prediction and highlight the potential for deploying multimodal, real-time monitoring systems on TinyML-compatible hardware. Nonetheless, challenges remain, particularly the scarcity of annotated swarm-specific datasets, variability in hive acoustics across regions, and limitations of on-device processing power. Future work should focus on sensor fusion, few-shot or continual learning to improve adaptability, and explainable AI tools to increase the interpretability and trustworthiness of swarm alerts in field conditions.

Section 3 has examined diverse TinyML applications that contribute directly to sustainable and efficient beekeeping practices, including hive condition monitoring, bee activity recognition, pest and disease detection, and predictive modeling for swarming events. Each application area emphasizes the use of lightweight, real-time inference models deployed directly at the hive, highlighting the advantages of edge computing in remote or resource-limited apiaries. Despite these promising developments, the success of TinyML solutions depends heavily on the availability and quality of relevant datasets, as well as consistent benchmarks for evaluating model accuracy and efficiency on resource-constrained hardware. To address these critical considerations, the following section (Section 4) reviews existing datasets and benchmarking practices, exploring their roles in facilitating further advancements in TinyML-driven apicultural systems.

## 4 Datasets and Benchmarking for TinyML in Beekeeping

Effective TinyML applications rely significantly on high-quality datasets and standardized benchmarking methods to train, evaluate, and compare models. Datasets provide the foundation for training and validating ML models, while benchmarking enables the evaluation of these models under constraints typical of edge devices, such as limited memory, processing power, and energy resources. This section explores existing datasets in the beekeeping domain, discusses current benchmarking metrics and tools applicable to TinyML, and highlights the limitations and gaps that hinder widespread adoption and scalability of these technologies.

## 4.1 Existing Datasets for TinyML in Beekeeping

The successful implementation and evaluation of TinyML applications in beekeeping depend critically on the availability of high-quality, well-labeled datasets collected under realistic hive conditions [12, 136]. Public datasets now span multiple modalities, including acoustic recordings, environmental sensor streams (e.g., temperature, humidity, pressure, $CO_2$), and visual data (images or videos) from hive entrances or colony inspections. However, significant limitations persist: many corpora remain small, inconsistently labeled, or biased toward narrow ecological conditions. Datasets designed explicitly for TinyML, such as those with low-resolution inputs, compact memory footprints, and annotations suitable for on-device inference, are still rare. Consequently, researchers often resort to adapting general-purpose datasets or developing custom corpora tailored to their embedded deployment scenarios [60].

A range of datasets has emerged to address these needs (Table 2). Early acoustic benchmarks such as NU-Hive, BUZZ, OSBH, and To Bee or Not to Bee provide foundational audio resources labeled for colony states and activity levels. More comprehensive multimodal collections such as MSPB integrate long-term audio features with temperature and humidity, while the Smart Bee Colony Monitor dataset couples hive environmental streams with annotated clips suitable for queen state analysis. The UrBAN dataset represents a major advance, capturing over 3,000 hours of audio synchronized with inspection-level phenotypic metadata (e.g., population, queen status, Varroa presence), thereby enabling supervised learning in real urban conditions [5].

Vision-based corpora are also expanding. Kulyukin-BeeVid offers YOLO-formatted bounding boxes from entrance cameras to benchmark lightweight detectors, while the Varroa Dataset and BeeImages Dataset target health classification tasks such as Varroa infection, robbing, and missing queen states [50]. The Pollen Dataset provides a small but widely cited image benchmark for distinguishing pollen-bearing bees [106]. More recently, the VnPollenBee dataset has been introduced as the first large-scale, fully annotated resource for pollen-bearing bee detection. Comprising 2051 images and over 60,000 bounding boxes (1,758 pollen-bearing vs. 59,068 non-pollen-bearing), VnPollenBee directly addresses the real-world imbalance challenge at hive entrances and offers a strong benchmark for evaluating detection models under TinyML constraints [78]. Collectively, these resources mark a growing but still fragmented ecosystem of datasets, underscoring the need for standardized, TinyML-oriented corpora to advance reproducible and scalable research in precision apiculture.

Table 2. Overview of publicly available beekeeping datasets categorized by hives, labels, modalities, size, and accessibility.

| Name | Hives | Labels | Modality | Size | Availability |
|------|-------|--------|----------|------|--------------|
| NU-Hive [24] | 2 | Queenright/queenless | Raw audio<br>Temperature<br>Humidity | 96 h<br>—<br>— | Public<br>Not available<br>Not available |
| BUZZ [59] | 6 | Bee buzzing/cricket/noise | Raw audio | 10,260 and 12,914 labeled audio samples | Public |
| OSBH [83] | 6 | Queenright/queenless | Raw audio | 140 min | Subset only |
| Too bee or not to bee [83] | 8 | Bee buzzing/no bee buzzing | Raw audio | 12 h | Public |
| MSPB [144] | 53 | Population; Honey yield; Queen conditions; Hygienic behavior; Winter mortality | Hand crafted audio features<br>Temperature<br>Humidity | Quarter-hourly during 365 day | Public |
| Smart Bee Colony Monitor [48] | 4 | Queenright/queenless | Raw audio<br>Temperature<br>Humidity<br>Pressure | 118 h<br>Hourly during 40 days<br>Hourly during 40 days<br>Hourly during 40 days | Public |
| UrBAN [5] | 10 | Population; Queen conditions;Winter mortality | Raw audio<br>Temperature<br>Humidity | 3171 h<br>Quarter-hourly during 135 day | Public |
| Kulyukin-BeeVid [60] | 4 | Flying bee bounding boxes | Images (from video), YOLO-format annotations | 23,173 labeled bees in 5,819 frames | Public |
| BeeTogether [20] | - | Queen presence/ No queen | Raw audio | 13h | Public |
| Varroa Dataset [50] | - | Healthy, Infected | RGB Images | 13,507 Images | Public |
| BeeImages Dataset [50] | - | Ant problems, Varroa, Missing queen, Robbing, etc. (6 classes) | RGB Images | 5,172 images | Public |
| Pollen Dataset [106] | - | Pollen / Non-pollen | RGB Images | 714 images | public |
| VnPollenBee [78] | - | Pollen / Non-pollen | RGB Images | 2051 images, 60,826 boxes | public |
| Ngo et al. [77] | 5 | Pollen / Non-pollen | RGB Images (video frames) | Custom dataset: 3,000 training + 500 testing images (640×480), manually annotated | Not public |

## 4.2 Benchmarking Methodologies and Practices

Benchmarking methodologies provide standardized metrics and evaluation criteria to assess TinyML models. These assessments typically consider model accuracy, efficiency, and resource consumption, which are particularly important for beekeeping applications where devices must operate under strict resource constraints. Although many datasets are available to support model training, only a limited number of studies explicitly report benchmarking practices that reflect TinyML conditions, such as low power operation, restricted memory, or on-device inference. The variation in evaluation practices also reflects the absence of a common standard across the field.

One of the most comprehensive efforts was conducted by Kulyukin et al.[60], who compared several YOLO variants for bee detection. Their evaluation included conventional metrics such as precision, recall, F1-score, and IoU, as well as custom measures such as energy efficacy and operational footprint, directly addressing the trade-off between accuracy and energy consumption in embedded environments. Abdollahi et al.[5] introduced the UrBAN dataset, which enables population prediction using multimodal sensor data and evaluated regression models with MAE, RMSE, and correlation metrics. BeeTogether [20] extended benchmarking to standardized acoustic datasets for queen presence classification, reporting accuracy and confusion matrices for both GMM and CNN models.

Other studies focused on benchmarking approaches using custom datasets. Schurischuster et al.[111] evaluated traditional classifiers including Naïve Bayes, SVM, and Random Forest for mite detection. Their study compared several feature extraction methods such as color histograms, histogram moments, local binary patterns, and SURF, with results reported in terms of accuracy and F1-score. Ngo et al.[77] developed an imaging system based on YOLOv3-tiny running on a Jetson TX2 device, achieving an F1-score of 0.94 for pollen recognition and mean absolute percentage errors of 8.45 percent for pollen counts and 10.55 percent for total counts, validated against manual counts and pollen traps.

In contrast, Zhu et al.[144] introduced the MSPB dataset, which offers longitudinal data but does not include benchmarking results. Earlier acoustic studies also contributed useful baselines. Nolasco and Benetos[83] reported binary classification accuracy for bee versus no-bee recognition using MFCC features. Cecchi et al.[24] applied FFT-based spectral analysis to characterize swarming signals, while Terenzi et al.[119] compared acoustic feature extraction techniques, including STFT, MFCC, and CWT, under different evaluation splits. Although these works did not focus on embedded deployment, they demonstrate careful benchmarking of feature extraction pipelines and provide valuable methodological insights.

Table 3 summarizes these contributions, showing the range of evaluation approaches and identifying gaps in current benchmarking practices for TinyML-based beekeeping applications.

## 4.3 Limitations and Gaps in Current Practices

Despite the growing number of datasets and benchmarking studies, several critical limitations hinder the maturity and scalability of TinyML in beekeeping. First, the majority of datasets remain fragmented and domain-specific, often restricted to a single sensing modality such as audio or vision. While multimodal resources like UrBAN and MSPB have begun to integrate acoustic, visual, and environmental data, standardized datasets explicitly designed for TinyML constraints, such as reduced resolution input, memory efficiency, and suitability for on-device inference, are still scarce. Furthermore, many publicly available corpora are limited in size, poorly balanced across classes, or confined to narrow ecological conditions, making it difficult to generalize models to diverse apiary contexts.

Second, benchmarking practices lack consistency across studies. Researchers frequently report conventional metrics such as accuracy or F1 score without addressing TinyML-specific considerations like latency, memory footprint, energy consumption, and long-term operational stability. Only a few studies, such as Kulyukin et al. [60], explicitly evaluate trade-offs between accuracy and energy use on embedded devices, while most others rely on desktop-class evaluations that do not reflect field conditions. This inconsistency prevents meaningful comparison of results across different approaches and limits reproducibility.

Third, the absence of unified benchmarking frameworks tailored to beekeeping hampers progress. Unlike the broader TinyML community, where initiatives such as MLPerf and MLPerf Tiny provide standardized workloads, accuracy targets, and scenario-specific metrics for fair comparison [68, 101], apiculture research lacks a comparable reference suite. Current studies typically resemble the open division of MLPerf, with custom models and datasets, but without the closed division needed to enable apples-to-apples comparison. This gap results in heterogeneous evaluation pipelines and fragmented reporting, slowing the translation of promising methods into deployable systems.

Table 3. Overview of benchmarking studies on AI-based beehive monitoring.

| Source | Models Tested | Metrics Reported | Hardware | Benchmarking Focus |
|---|---|---|---|---|
| Kulyukin et al. [60] | YOLOv3, YOLOv4-tiny, YOLOv7-tiny | Precision, Recall, F1, IOU, Energy Efficacy, Operational Footprint | GTX-980 GPU, Raspberry Pi | Accuracy vs. energy trade-off in bee object detection |
| UrBAN [5] | CNN, Random Forest, Regression | MAE, RMSE, MAPE, Pearson correlation, $R^2$ | Python pipeline with feature extraction | Population prediction via multimodal sensor data |
| BeeTogether [20] | GMM, CNN, MFCC-based classifiers | Accuracy, Confusion Matrix | Not specified | Queen presence detection from standardized audio |
| MSPB [144] | Not reported | None (dataset only; enables downstream ML) | – | Trait estimation (population, queen, honey) |
| Nolasco et al. [83] | SVM, CNN | Accuracy (Bee / NoBee), MFCC-based input | Desktop | Beehive presence classification via sound |
| Cecchi et al. [24] | None (FFT + spectral analysis) | Frequency range visualization, swarm signal observation | Custom audio system | Preliminary acoustic feature analysis for hive state detection |
| Schurischuster et al. [111] | Naïve Bayes, SVM, Random Forest | Accuracy, F1-score, 3-fold CV | Python (OpenCV, Scikit-learn), Desktop | Comparative evaluation of feature extraction (color histograms, histogram moments, LBP, SURF) and classifiers for Varroa detection |
| Ngo et al. [77] | YOLOv3-tiny + Kalman filter | Accuracy, MAE | Jetson TX2 GPU (edge device) | Benchmarking lightweight object detection for pollen-bearing bee classification |

Finally, there are practical challenges related to data annotation and accessibility. Many datasets rely on manual labelling of images or audio, which is time-intensive and prone to subjectivity. Others remain inaccessible due to privacy restrictions or limited sharing policies, forcing researchers to depend on custom corpora that cannot be reproduced by others. Moreover, scenario-based evaluations, such as distinguishing between low-latency single-stream inference at hive entrances and offline batch analysis for population estimation, remain unexplored in current work. These issues collectively highlight the need for community-driven efforts to establish standardized TinyMLready datasets, along with comprehensive benchmarking frameworks that incorporate both model accuracy and edge device performance metrics. Addressing these gaps will be essential to scale TinyML solutions for robust, field-ready beekeeping applications.

## 5  Challenges and Future Directions

Despite significant progress in applying TinyML to beekeeping, several challenges persist that limit the scalability and robustness of current solutions. These challenges broadly relate to hardware constraints, model optimization trade-offs, data availability, deployment barriers, and ecological and ethical considerations. Addressing these issues will be crucial for advancing the practical adoption of TinyML in apiculture.

### 5.1  Hardware Constraints

The effectiveness of TinyML applications in beekeeping is tightly coupled with the capabilities and limitations of the underlying hardware. Available platforms span microcontrollers (MCUs) such as the Arduino Nano 33 BLE Sense and STM32 boards, single-board computers (SBCs) such as the Raspberry Pi and Jetson Nano, dedicated AI accelerators including the Coral Edge TPU and Kendryte K210, and emerging ultra-low-power (ULP) chips such as the Syntiant NDP101 and GAP9 [44, 63, 99]. Each category offers distinct trade-offs. MCUs are inexpensive and energy-efficient, making them attractive for acoustic or vibration sensing, but their kilobyte-scale memory prevents deployment of high-dimensional models. SBCs can run complex convolutional networks for bee detection and *Varroa* screening, but at the cost of high power draw, which limits their use in battery-only apiaries [5, 60]. Accelerators such as the Coral Edge TPU provide a middle ground by enabling fast inference per watt, though portability across model architectures and toolchains remains limited. Meanwhile, ULP chips are promising for continuous, always-on acoustic monitoring, but are still immature for real deployments.

Table 4 summarizes these categories in terms of strengths, limitations, and demonstrated beekeeping applications. To complement this, Figure 6 provides a radar-style visualization of relative trade-offs across five dimensions: accuracy/model capacity, energy efficiency, memory/compute headroom, cost, and field suitability. The scores are synthesized from published specifications and representative deployments rather than direct benchmarks, and are therefore intended as a conceptual map of trade-offs rather than absolute measurements. Together, the table and figure illustrate a fundamental challenge, where no single hardware class fully satisfies the competing requirements of high accuracy, low energy consumption, and low cost in real-world apiary settings. Future progress will depend on co-design strategies that optimize both hardware and software. For example, lightweight ML frameworks such as TensorFlow Lite Micro or Edge Impulse can be tuned to exploit device-level features, while specialized accelerators and neuromorphic processors offer the potential to reduce the accuracy–energy trade-off. Equally important is the development of standardized benchmarks that measure accuracy, latency, memory, and power consumption under realistic hive-monitoring conditions, enabling more rigorous comparison of hardware platforms for apicultural TinyML deployments.

Addressing these constraints requires continued innovation in both hardware and software design. Future research should explore ultra-low-power model architectures, event-driven inference strategies, efficient data compression techniques, and hardware-aware neural architecture search (NAS) tailored for beekeeping use cases [107, 113]. Integrating energy-harvesting methods and deploying custom AI accelerators, such as the Syntiant NDP101 or Google Edge TPU may also enhance the feasibility and scalability of TinyML solutions in harsh field conditions [114]. Ultimately, overcoming these hardware challenges is essential for the successful and sustainable deployment of intelligent monitoring systems in modern apiculture.

### 5.2    Model Optimization Trade-offs

Model optimization is central to deploying TinyML in beekeeping, where devices must balance predictive accuracy with tight computational and energy budgets. Common techniques include quantization, pruning, and knowledge distillation, each aiming to compress models while preserving acceptable performance. Quantization, for example, can reduce memory footprint and latency substantially, but in tasks such as Varroa mite detection, it may blur subtle visual cues critical for reliable classification [60]. Pruning removes redundant weights to accelerate inference and reduce storage demands, yet aggressive pruning may degrade accuracy, especially in small, imbalanced datasets. Knowledge distillation offers an alternative by transferring knowledge from a large, accurate teacher model into a smaller student model, though this introduces additional training complexity[113].

Table 4. Comparative overview of hardware platforms for TinyML in beekeeping.

| Category | Representative Devices | Strengths | Limitations | Beekeeping Relevance / Examples |
|---|---|---|---|---|
| MCUs | Arduino Nano 33 BLE Sense; STM32; ESP32 | Ultra–low power; low cost; simple deployment; good for always-on sensing | KB-level RAM/flash; limited compute; unsuitable for high-dimensional inputs (images, long spectrograms) | Acoustic/vibration sensing for swarm cues or activity level; on-device classical ML or tiny CNNs [12, 136] |
| SBCs | Raspberry Pi 4/5; NVIDIA Jetson Nano/Orin Nano | High compute; GPU support; runs full CNNs (YOLO, EfficientNet) | Higher power draw; thermal management; less ideal for battery/solar only sites | Bee detection/counting and Varroa screening with YOLO-tiny on Raspberry Pi/Jetson [5, 60] |
| Accelerators for AI | Google Coral Edge TPU; Kendryte K210 | High inference/W; hardware offload; low latency at the edge | Model portability limits; conversion/toolchain quirks; extra BOM cost | Prototype CNN inference for entrance monitoring; efficient image pipelines when power-constrained [12, 63] |
| Emerging ULP Chips | Syntiant NDP101; GreenWaves GAP9; RISC-V ML cores | Sub-mW alwayson; designed for audio/keyword-style tasks; promising efficiency | Early ecosystem; limited availability; narrow model classes | Continuous acoustic swarm monitoring; wake-word style hive-state triggers for duty-cycled vision [44, 99] |

Another difficulty lies in the absence of standardized benchmarks for evaluating optimized models. Most studies report conventional metrics such as accuracy, F1-score, or recall, while ignoring TinyML-specific constraints, including latency, memory footprint, or energy per inference. This lack of uniform evaluation makes it challenging to compare models across modalities (acoustic, visual, environmental) or hardware platforms, hindering reproducibility and progress [5, 136].

The study by De Nart et al. [28] further illustrates the optimization dilemma. While large CNNs such as Inception-ResNet V2 achieved nearly perfect subspecies classification, their computational complexity makes them unsuitable for TinyML deployment in field hives. In contrast, MobileNetV2, a lightweight architecture with approximately 3.4 million parameters, delivered competitive accuracy ( 0.93), suggesting that compact models can provide a feasible balance between efficiency and predictive power. This reinforces the broader challenge in apicultural TinyML: optimization strategies must selectively trade accuracy for efficiency to enable reliable on-device inference under resource constraints.

Recent work has begun to explore co-optimization frameworks that jointly consider model architecture, compression strategy, and device-level constraints. Hardware-aware neural architecture search (NAS) has been proposed to automatically design efficient architectures tailored for edge deployment. Unlike post hoc techniques such as pruning or quantization, which compress existing models at the risk of accuracy loss, NAS can generate architectures that are explicitly optimized for a given hardware budget, balancing accuracy, memory footprint, latency, and energy consumption. By making efficiency an objective of the search process, NAS has the potential to yield task-specific lightweight models that outperform generic architectures such as MobileNet or YOLO-tiny under field constraints [107]. Similarly, mixed-precision quantization, structured pruning, and lightweight backbones such as MobileNetV2, EfficientNet-Lite, or YOLO-tiny variants provide promising trade-offs between performance and resource usage. Nevertheless, their systematic application in beekeeping remains underexplored, with most studies adapting general-purpose models rather than developing domain-specific pipelines [99].

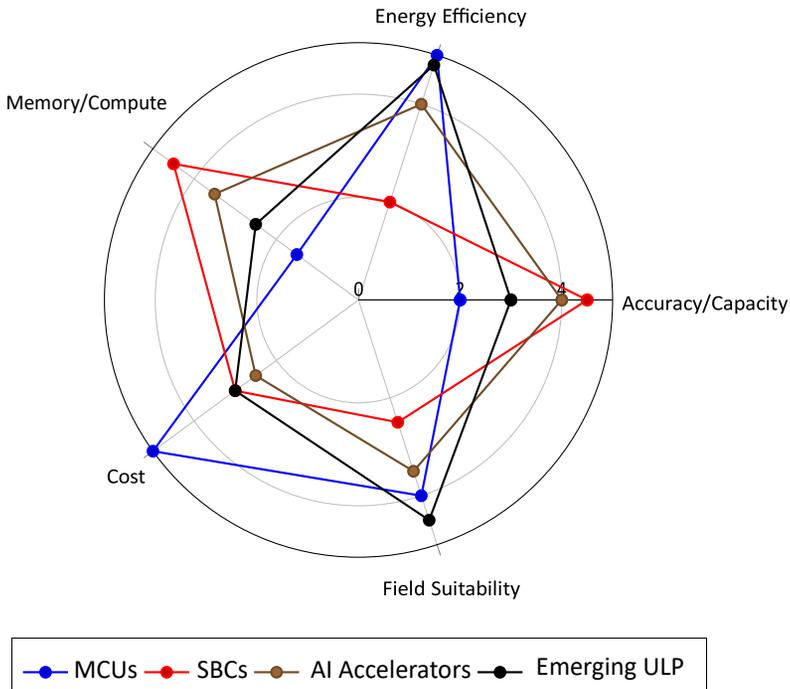

Fig. 6. Radar comparison of hardware categories for TinyML in beekeeping (scores 1–5; higher is better).

Table 5 summarizes machine learning models applied in beekeeping across different sensing modalities, highlighting their advantages, limitations, and representative studies. These examples underscore both the potential and the trade-offs involved in compressing and optimizing models for field-ready TinyML deployment. Future research should focus on developing open-source benchmarking frameworks that explicitly evaluate accuracy–energy trade-offs, robustness under environmental variability, and long-term reliability, ensuring optimization strategies translate from laboratory prototypes to sustainable apiary deployments.

## 5.3    Data Challenges

Despite recent progress in dataset development, significant challenges remain that hinder the scalability of TinyML in beekeeping. As summarized in Table 2, current resources are fragmented across modalities, with most efforts concentrated in acoustic corpora and far fewer datasets covering visual, environmental, or multimodal signals. Figure 7 highlights this imbalance: audio datasets dominate but are often limited to narrow labels such as queenright/queenless or buzzing/no buzzing, visual corpora are confined to relatively small image collections for tasks like pollen or *Varroa* detection, and multimodal resources are almost absent, with UrBAN representing the only comprehensive example. This uneven distribution restricts the generalizability of trained models across diverse hive contexts, reduces the reliability of benchmarking studies, and limits opportunities for robust multimodal fusion approaches [62]. Furthermore, many datasets remain small in scale, class-imbalanced, or partially inaccessible, constraining reproducibility and slowing the transition from laboratory prototypes to field-ready TinyML deployments.

## 5.4    Deployment and Scalability

Deployment of TinyML in beekeeping is constrained by energy budgets, intermittent connectivity, and limited on-device resources. Studies demonstrate that edge-based inference reduces bandwidth demands and enables timely alerts, especially when paired with solar-powered MCUs or Raspberry Pi gateways. Optimizations such as quantization and lightweight backbones (e.g., YOLOv7-tiny, int8 audio classifiers) have shown favorable trade-offs between accuracy and energy footprint, confirming the practicality of real-time vision and acoustic monitoring in field hives [74, 81].

Scalability depends heavily on communication and architecture design. Low-power wide-area networks (e.g., LoRaWAN) have proven effective for multi-hive deployments, where only compact telemetry or event flags are transmitted to conserve energy [102]. Tiered designs are common: edge nodes handle inference, hive-level gateways fuse sensor data, and cloud layers aggregate summaries for long-term analytics and remote updates. Platforms such as Beemon and BeeLive illustrate this integration, while emphasizing modularity, containerization, and maintainability for distributed apiaries [40, 104, 117]

Table 5. Machine learning models applied in beekeeping across different sensing modalities.

| Modality | Models Used | Advantages | Limitations | Example Studies |
|---|---|---|---|---|
| Acoustic / Vibrational | Naïve Bayes, KNN, SVM, CNN, LSTM, Transformer | Detect swarm states; CNN/LSTM reach ≥97% accuracy; temporal models capture time dynamics | Require large labeled datasets; sensitive to background noise; MCU deployment challenging | Iqbal et al. [43] (NU-hive CNN 99%, SVM 97%); Bencsik et al. [16] (vibrational swarming patterns) |
| Visual | YOLOv3, YOLOv4tiny, YOLOv7tiny, Inceptionv3, MobileNetV2, EfficientNet-Lite | Detect bees, mites, brood stages; YOLOtiny/MobileNet feasible on edge devices | High compute/memory demand; accuracy–energy trade-offs; sensitive to lighting/occlusion | Kulyukin et al. [60] Rathore & Agrawal [96] (brood classification, 97%) |
| Environmental Sensors | Regression models (Linear, SVR), Decision Trees, Random Forest, shallow CNNs | Predict colony strength, population trends, hive health from multimodal inputs (temp, humidity, $CO_2$) | Limited predictive power vs. deep models; sensitive to seasonal/environmental variation | Abdollahi et al. [5] (UrBAN multimodal dataset regression for population prediction) |
| Chemical VOC-/ based | PCA + ML classifiers, CNNs on sensor outputs | Disease detection (foulbrood, mites) with lowcost gas sensors; PCA reveals VOC patterns | Datasets small; VOC signals overlap with environment; little TinyML field deployment | Rigakis [102] (VOC Monitoring); Szczurek [115] Zhao [143] (VOC-based disease signatures) |
| Integrated / Multi-modal | Autoencoders, Fusion CNNs, Hybrid ML pipelines | Combine acoustic, visual, environmental cues; autoencoders detect anomalies; robustness to sensor failure | High complexity; increased hardware demands; no standard fusion benchmarks | Machhamer et al. [64] (IoT anomaly detection); Abdollahi et al. [5] (multimodal UrBAN) |

Standardization and dataset diversity remain critical for deployment at scale. Frameworks like MLPerf Inference: Tiny offer reproducible metrics for accuracy, latency, and energy use, and could be adopted for hive-monitoring benchmarks. [101] Multimodal datasets such as UrBAN stress-test models across diverse conditions and seasons, highlighting the need for evaluation campaigns beyond single-site trials. Collectively, the literature suggests a deployment checklist: energyfirst hardware selection, compact model design, event-driven sensing, LPWAN communication, standardized benchmarking, and edge-focused MLOps, ensuring TinyML systems for apiculture can move from prototypes to reliable, fleet-scale solutions [108, 120].

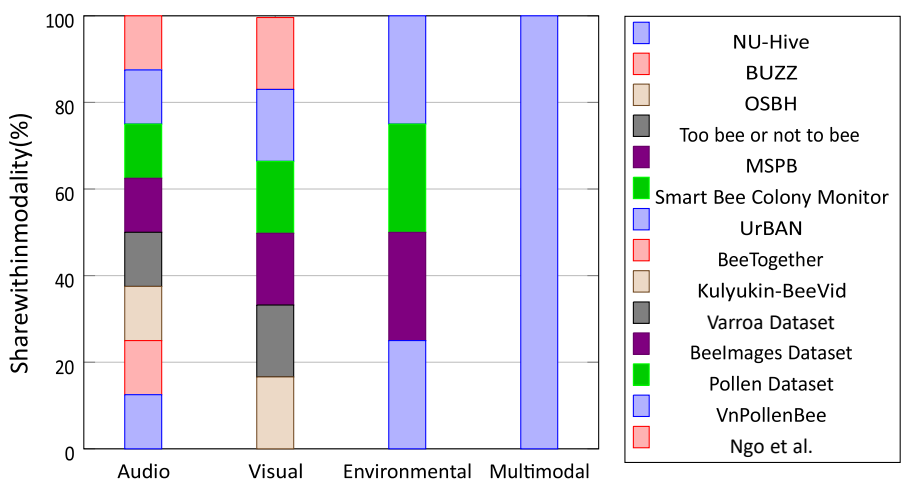

Fig. 7. Per-modality dataset composition (percent shares within each modality) based on datasets listed in Table 2. Bars sum to 100% per modality, highlighting the scarcity of multimodal resources and the uneven distribution across audio, visual, and environmental datasets.

## 6　Discussions and Ecological Impact

The integration of TinyML and IoT in beekeeping extends beyond technical advancements to broader ecological and societal implications. While these technologies offer promising solutions for hive health monitoring, early pest detection, and improved management practices, their adoption raises questions about long-term sustainability, accessibility, and ecological balance. The deployment of lightweight models on edge devices reduces reliance on invasive inspections and centralized infrastructures, potentially lowering stress on bee colonies and minimizing carbon footprints associated with data transmission. At the same time, considerations around technological benefits, ecological trade-offs, stakeholder engagement, and ethical responsibilities are critical to ensure that innovation in precision apiculture supports the delicate equilibrium of pollinator ecosystems.

### 6.1　Technological Benefits

In Australia, beekeeping contributes not only to honey production but also to vital pollination services for high-value crops such as almonds, apples, and canola [118]. The integration of TinyML into apiculture offers substantial benefits by enabling localized, on-device processing of sensor data. This reduces dependence on cloud-based infrastructures and continuous network connectivity, which is particularly advantageous for remote and semi-arid apiary sites where power and internet access remain limited.

Edge-based TinyML systems further support real-time analysis of acoustic, thermal, and motion signals, enabling early detection of hive stressors such as queen loss, disease outbreaks, or abnormal foraging activity. These non-invasive and continuous monitoring capabilities also facilitate strategic breeding programs by capturing phenotypic traits relevant to productivity and resilience, including honey yield and Varroa resistance [135].

A practical example of this approach is the BeeStar platform, developed and tested under Australian field conditions. By leveraging microcontroller-based hardware optimized for low energy consumption, BeeStar demonstrates the feasibility of deploying TinyML for real-time hive monitoring at scale [14]. These innovations not only advance precision apiculture but also align with broader national priorities such as the Ag2030 Plan, which seeks to expand Australia's agricultural sector to AUD 100 billion by 2030 through digital transformation and adoption of smart technologies [30].

## 6.2 Ecological Considerations

Australia's environment supports over 1,700 native bee species and diverse pollination ecosystems, making ecological sensitivity a critical factor in the adoption of IoT and TinyML for beekeeping. Compared to conventional hive inspections, passive sensing technologies offer a less disruptive means of data collection, reducing colony stress and supporting broader biodiversity conservation. Nevertheless, the ecological footprint of deploying electronic devices must be carefully managed, particularly with respect to e-waste, bushfire resilience, and material sustainability. Integrating solarpowered units, weather-resistant housings, and recyclable components can mitigate these risks and ensure longer system lifespans under harsh field conditions. Furthermore, the National Honey Bee Breeding Strategy 2024–2029, which has established reference sites across New South Wales, Queensland, and Western Australia, provides a valuable framework where sensor-equipped hives can be used to monitor phenotypic traits and environmental interactions in both Varroa-affected and unaffected regions [88, 92].

## 6.3 Stakeholder Implications

Australia's apiculture sector consists of commercial operators, semi-commercial enterprises, and hobbyist beekeepers, each with distinct needs and resources. For small to mid-sized operators, access to affordable and low-maintenance tools is critical, as technical expertise and infrastructure are often limited. TinyML-enabled hive monitoring systems that are simple to install and operate can democratize access to precision apiculture technologies, supporting both productivity and resilience. The AgriFutures Honey Bee & Pollination Program has emphasized innovation, bee health, and industry sustainability in its 2020–2025 RD&E Plan, providing a policy framework that aligns with the integration of edge-based AI monitoring [8]. Similarly, the National Honey Bee Breeding Strategy 2024–2029 underscores the importance of standardized phenotypic data collection to accelerate selective breeding, particularly for Varroa resistance, a national priority since the 2022 incursion and the cessation of eradication efforts in 2023 [88]. TinyML platforms can contribute to this by enabling consistent field-level data capture across diverse apiary conditions. Institutions such as Tocal College and the University of New England (AGBU), which already provide infrastructure for genetic evaluation and industry training, represent key stakeholders for piloting and scaling smart hive systems that bridge research, education, and practice [26].

## 6.4 Policy and Ethics

The deployment of TinyML-enabled monitoring systems in beekeeping navigates a complex web of ethical and regulatory considerations, particularly in the Australian context. On the policy side, adherence to established frameworks such as the Australian Honey Bee Industry Biosecurity Code of Practice is paramount. This national code mandates rigorous biosecurity planning, disease surveillance, and record-keeping, serving as a blueprint for integrating advanced sensor systems without compromising established protocols [1]. Ethically, passive IoT monitoring methods, when thoughtfully designed, can reduce hive disturbance and chemical interventions, aligning with principles of humane and sustainable apiculture. However, beyond the immediate welfare of bees, broader concerns arise around data governance and the sovereignty of rural stakeholders. Drawing from ethical frameworks like "Responsible Innovation" and emerging discourse in biosurveillance ethics, it is crucial to ensure that data collected from hive-monitoring technologies respects the autonomy of beekeepers, clarifies ownership rights, and is implemented transparently via stakeholder-inclusive co-design processes [31].

Deploying AI-based systems in agriculture necessitates a careful consideration of data ethics, transparency, and farmer autonomy. The Australian Government's AI Ethics Principles provide a framework for ensuring technologies remain fair, inclusive, and environmentally sustainable. Beekeepers should retain ownership of their hive data, and monitoring systems must prioritize explainability and compatibility with traditional practices. Given the intersection between agriculture, Indigenous stewardship, and ecological conservation, new technologies must be designed to complement existing knowledge systems. Policies supporting open standards, training, and infrastructure development will be crucial to achieving equitable adoption of smart beekeeping innovations across Australia.

## 7    Conclusion

In conclusion, the integration of TinyML and IoT in beekeeping presents a transformative opportunity to advance hive monitoring and management by enabling real-time, low-cost, and non-invasive detection of colony health indicators. While current research has demonstrated promising results across modalities such as acoustic, visual, and environmental sensing, challenges remain in terms of dataset quality, hardware limitations, and benchmarking practices tailored to edge devices. Addressing these gaps through standardized datasets, energy-efficient model optimization, and scalable deployment frameworks will be critical for ensuring reliability and accessibility. Ultimately, the continued development of TinyML-driven solutions can contribute not only to technological innovation but also to ecological sustainability and the resilience of global apiculture.